\documentclass{article}
\usepackage{spconf,amsmath,graphicx,textcomp, adjustbox, multirow, textgreek}
\usepackage{array,booktabs}

\newcolumntype{L}[1]{>{\raggedright\let\newline\\\arraybackslash\hspace{0pt}}m{#1}}
\newcolumntype{C}[1]{>{\centering\let\newline\\\arraybackslash\hspace{0pt}}m{#1}}
\newcolumntype{R}[1]{>{\raggedleft\let\newline\\\arraybackslash\hspace{0pt}}m{#1}}

\usepackage{wrapfig}
\usepackage{amsfonts}

\usepackage[hang,flushmargin]{footmisc} 

\usepackage{setspace}

\newlength{\bibitemsep}
\newlength{\bibparskip}
\let\oldthebibliography\thebibliography
\renewcommand\thebibliography[1]{%
	\oldthebibliography{#1}%
	\setlength{\parskip}{\bibitemsep}%
	\setlength{\itemsep}{\bibparskip}%
}

\newcommand\figupmargin{-4mm}
\newcommand\figdownmargin{-5.2mm}

\newcommand\tablebetweenmargin{-1.5mm}
\newcommand\secupmargin{-2mm}

\title{GENERATION OF MULTIMODAL JUSTIFICATION USING VISUAL WORD CONSTRAINT MODEL FOR EXPLAINABLE COMPUTER-AIDED DIAGNOSIS}
\name{Hyebin Lee\textsuperscript{1}, Seong Tae Kim\textsuperscript{2}, and Yong Man Ro\textsuperscript{1*} \thanks{
		\textsuperscript{*}Corresponding author}
	}
\address{\textsuperscript{1}Image and Video Systems Lab, School of Electrical Engineering, KAIST, South Korea\\
\textsuperscript{2}Computer Aided Medical Procedures, Technical University of Munich, Germany}

\begin{document}
%
\maketitle

\begin{abstract}
The ambiguity of the decision-making process has been pointed out as the main obstacle to applying the deep learning-based method in a practical way in spite of its outstanding performance. Interpretability could guarantee the confidence of deep learning system, therefore it is particularly important in the medical field. In this study, a novel deep network is proposed to explain the diagnostic decision with visual pointing map and diagnostic sentence justifying result simultaneously. For the purpose of increasing the accuracy of sentence generation, a visual word constraint model is devised in training justification generator. To verify the proposed method, comparative experiments were conducted on the problem of the diagnosis of breast masses. Experimental results demonstrated that the proposed deep network could explain diagnosis more accurately with various textual justifications.
\end{abstract}

\vspace*{\secupmargin}
\section{Introduction}
\label{sec:intro}
\begin{figure*}
    \centerline{\includegraphics[width=0.85\linewidth]{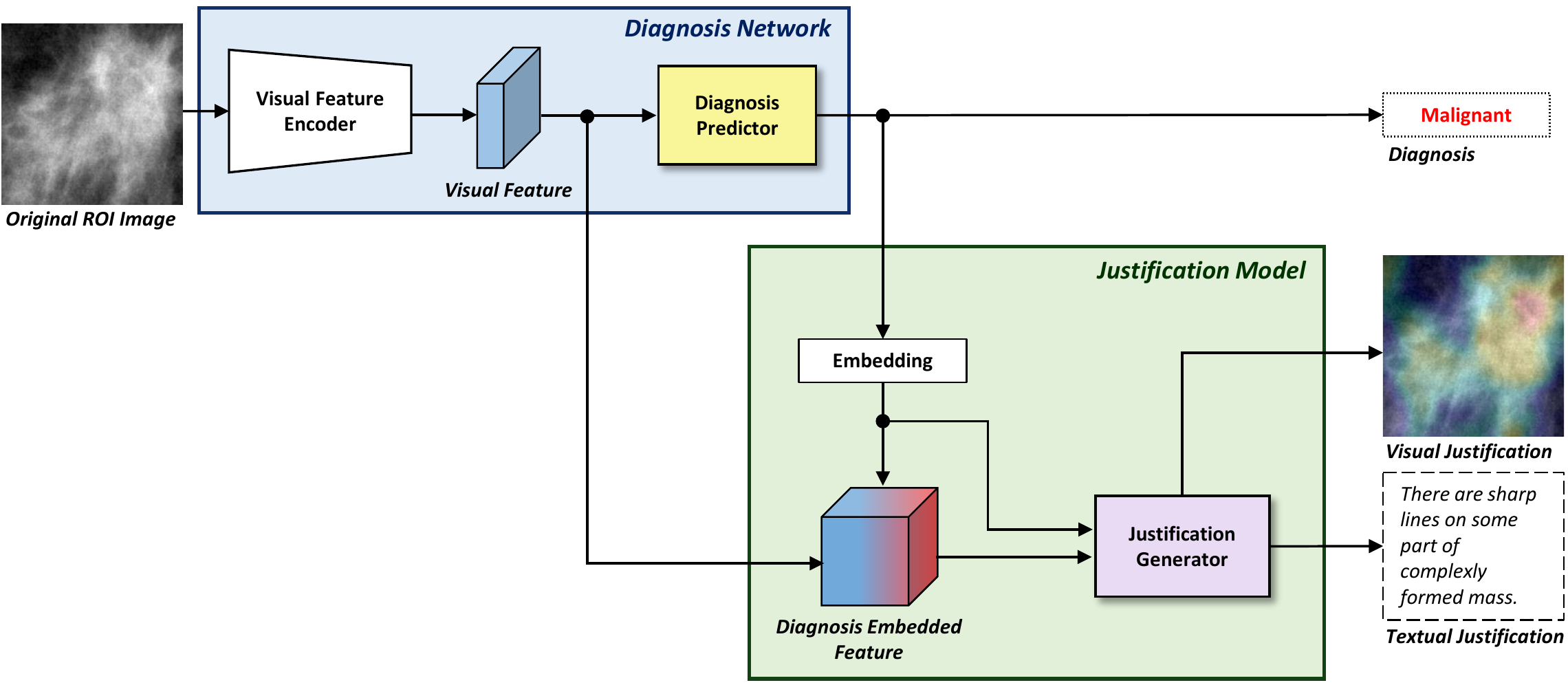}}
    \vspace*{\figupmargin}
    \caption{Overall proposed deep network framework for producing textual justification and visual justification.}
    \vspace*{\figdownmargin}
    \label{fig:overall}
\end{figure*}
Thanks to the remarkable achievements of deep learning technology, there are various attempts to utilize deep learning technique in many research fields such as image recognition \cite{krizhevsky2012imagenet, he2016deep} and medical image analysis \cite{ronneberger2015u, havaei2017brain}. Computer-aided detection (CADe) and Computer-aided diagnosis (CADx) also show notable successes with deep learning based approaches \cite{rajpurkar2017chexnet,kim2018icadx}.  On the contrary, difficulty in understanding the cause of a decision still remain as a dominant limitation for application of deep learning based method in the real world. To cope with this problem, several research efforts \cite{selvaraju2017grad,Fong_2017_ICCV, huk2018multimodal} have been devoted to developing the method for interpreting the decision of the deep network in recent years . \cite{selvaraju2017grad,Fong_2017_ICCV}  found the specific area of the input image which has the biggest impact on the final result. 
The multimodal approach \cite{huk2018multimodal} reported to generate explanation supporting the decision of deep network in form of attentive pointing map and text. 

In medical applications such as CADx, interpretation method reflecting reliability is more important, because it is mainly used in a high-risk environment directly connected to human health. There are several works which utilize prescribed annotation or medical report attached to the medical image as additional information for the decision explanation. \cite{kim2018visually, kim2019visual}  introduced critic network which exploits pre-defined medical lexicon to elaborate visual evidence of diagnosis. \cite{wang2018tienet,mdnet} proposed networks generate natural medical report from various Recurrent Neural Networks (RNNs) structure and point informative area of input medical image.

However, it is challenging to generate an accurate sentence with large variation because of the high complexity in the natural language. As addressed in \cite{Liu_2018_ECCV}, the conventional captioning methods suffer a problem in which the model duplicates a completely identical sentence of the training set even the model is trained on the large dataset. In other words, the deep network tends to memorize every sentence in the training set, which causes the situation that the generated sentence could not describe the target image in detail sufficiently. It becomes a more serious problem in the medical research area due to the limited number of medical report data.

In this study, we propose a novel deep network to provide visual and textual justification interpreting the diagnostic decision. The main contribution of this study is summarized as followings:

1) We propose a new justification generator to interpret the diagnostic decision of the deep network. The proposed justification generator could be constructed on top of the diagnosis network and provide the textual and the visual justification for the diagnostic decision. Due to the reason that the proposed justification generator is constructed on the diagnosis network, the proposed method could apply on any conventional CADx network (classifier of malignant mass and benign mass) to interpret the decision of the deep network without diagnostic performance degradation.

2) To overcome the duplication problem in which the model generates a completely identical sentence of the training set, we devise a new learning method with a visual word constraint loss. For evaluating the proposed method, a sentence dataset describing the characteristics (the shape and the margin) of breast masses with the words corresponding Breast Imaging Reporting and Data System (BI-RADS) mass lexicon has been collected in this study. Experimental results have shown that the proposed method could generate various textual justifications which are not just duplication of the sentence in the training dataset by guiding the textual justification generator with the visual word constraint model. 

The rest of the paper is organized as follows. In the Section \ref{sec:method}, we introduce the proposed diagnostic interpreting network for generating visual and textual justification. At the same time, we describe detail process to construct sentence dataset. Next, experimental results are presented and analyzed in terms of visual and textual justifying ability in Section \ref{sec:experiments}. Finally, Section \ref{sec:conclusion} concludes the paper.

\vspace*{\secupmargin}
\section{Proposed method}
\vspace*{\secupmargin}
\label{sec:method}

\begin{figure*}
    \centerline{\includegraphics[width=0.85\linewidth]{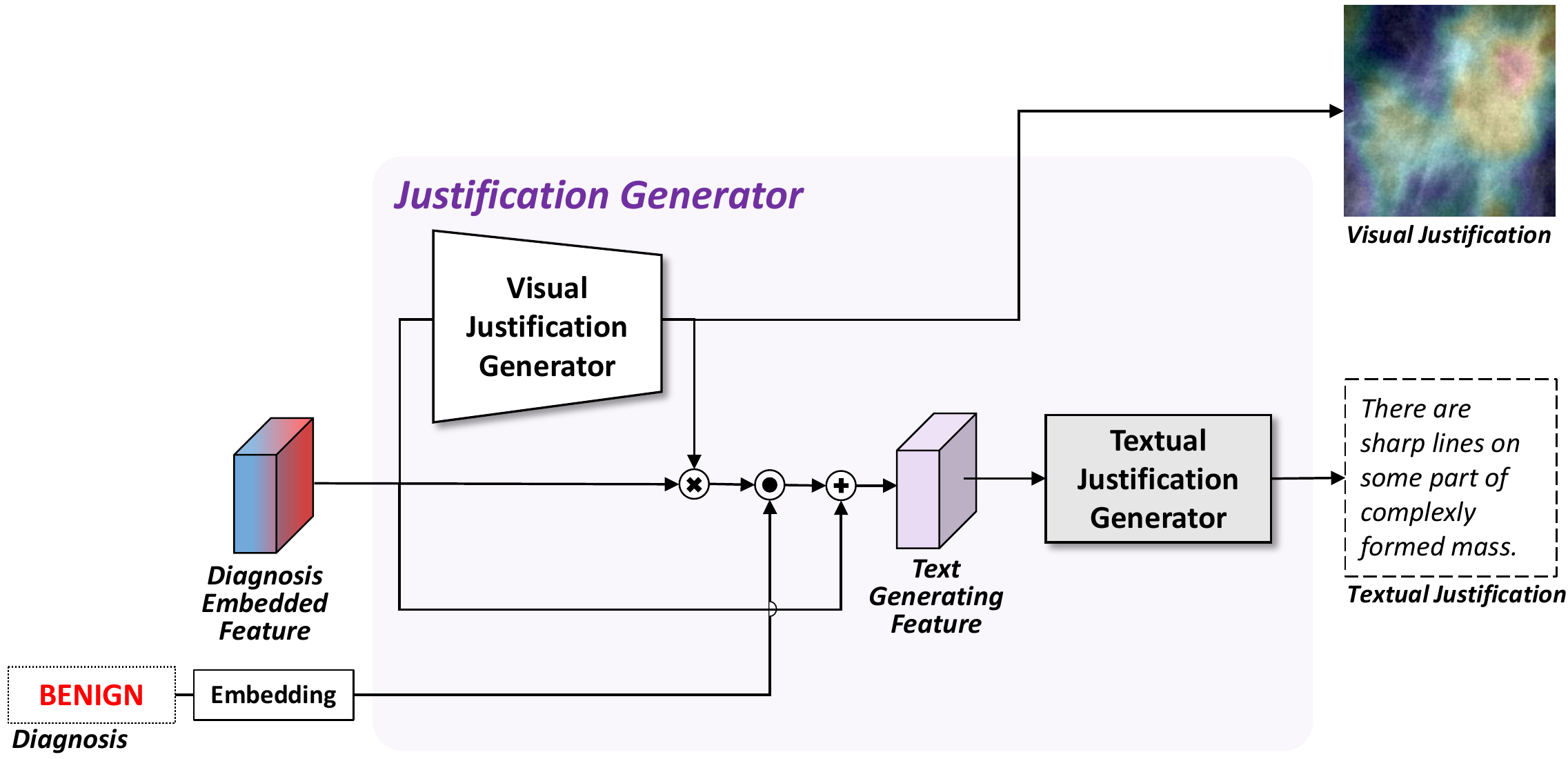}}
    \vspace*{\figupmargin}
    \caption{Detail architecture of the justification generator.}
    \vspace*{\figdownmargin}
    \label{fig:generator}
\end{figure*}

\begin{figure}
    \centerline{\includegraphics[width=0.95\linewidth]{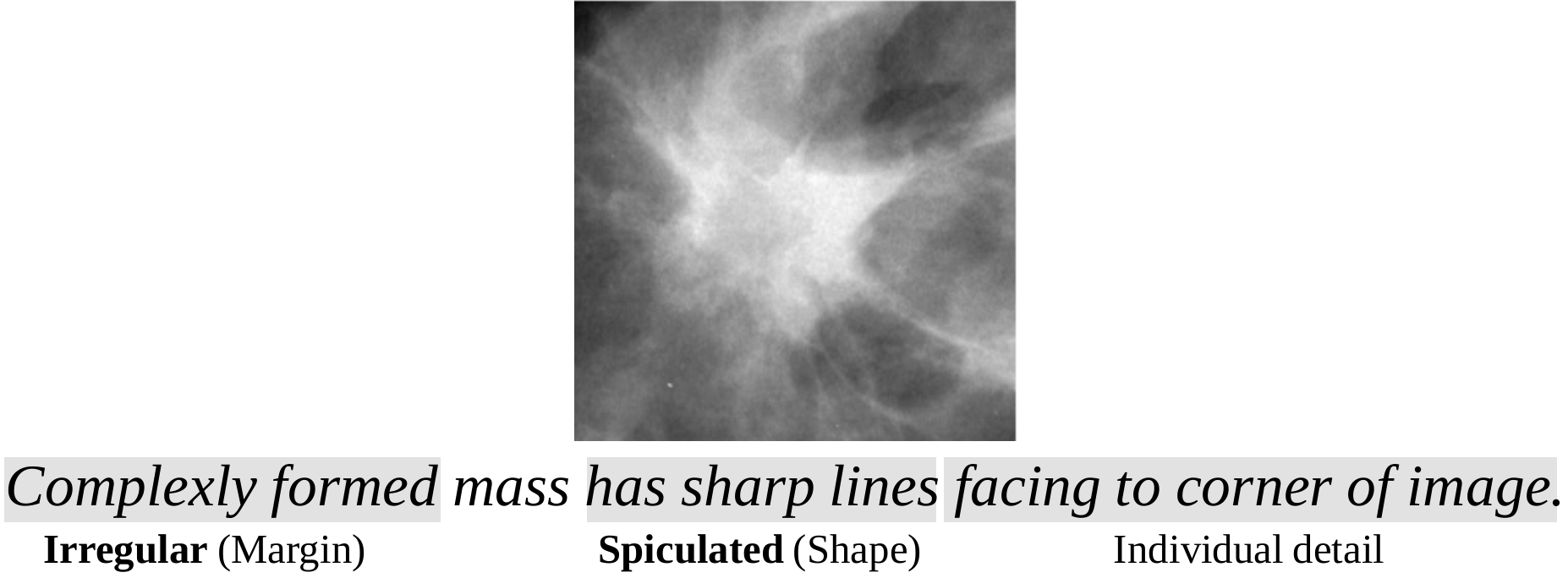}}
    \vspace*{\figupmargin}
    \caption{Example of composed sentence describing ROI mass image.}
    \vspace*{\figdownmargin}
    \label{fig:dataexample}
\end{figure}

\begin{figure*}
    \centerline{\includegraphics[width=0.95\linewidth]{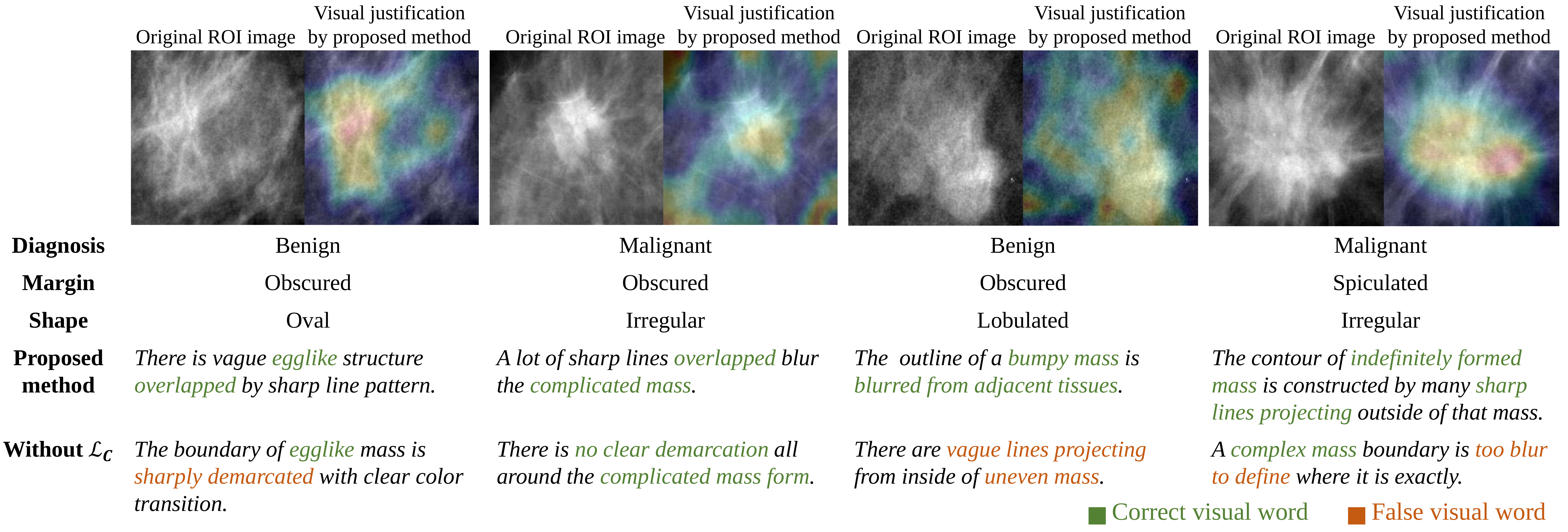}}
    \vspace*{\figupmargin}
    \caption{Results of the textual and the visual justification of the proposed method. Diagnosis, margin, and shape denote ground truth. The sentences for textual justification are compared with the proposed method and the method learned without \(\mathcal{L}_C \).}
    \vspace*{\figdownmargin}
    \label{fig:result1}
\end{figure*}

\subsection{Overall framework}
An overall proposed network framework is shown in Fig. \ref{fig:overall}. As shown in the figure, the overall architecture of the deep network is divided into two parts, a diagnosis network and a justification generator. As the diagnosis network, any conventional CADx network (classifier of malignant mass and benign mass) could be utilized. The justification generator employs a visual feature and a diagnostic decision of the diagnosis network. To effectively train the justification generator by avoiding the sentence duplication of the training set, a visual word constraint loss is devised in the training stage. The detailed structure of the justification generator and the learning strategy are described in following subsections.

\vspace*{\secupmargin}
\subsection{Justification generator}
As shown in Fig. \ref{fig:generator}, in order to explain the diagnostic decision, the justification generator make a textual justification and a visual justification from the predicted diagnosis and the visual feature. The visual feature is defined as an intermediate feature map in the diagnosis network. From given image \({\bf{I}}(n,m)\), the visual feature \({{\bf{f}}_v}(n,m,k)\) is extracted by the visual feature encoder \({{\rm{F}}_{{\varphi _{venc}}}}( \cdot )\) as
\begin{equation}
    {{\bf{f}}_v}(n,m,k) = {{\rm{F}}_{{\varphi _{venc}}}}({\bf{I}}(n,m)).
\end{equation}
The diagnosis predictor \({{\rm{D}}_{{\varphi _{dig}}}}\) make a diagnostic decision   \({{\bf{\hat y}}_d} = \{ {p_{benign}},{p_{malignant}}\} \) from the visual feature by
\begin{equation}
    {{\bf{\hat y}}_d} = {{\rm{D}}_{{\varphi _{dig}}}}({{\bf{f}}_v}),
\end{equation}
where \({p_{{benign}}}\) and \({p_{malignant}}\) denote the probability of the benign and the probability of the malignant, respectively. The diagnostic decision \({{\bf{\hat y}}_d}\) is embedded to the channel-wise attention weight \({\alpha _d}^{embed}\) as followings:
\begin{equation}
    \alpha _d^{embed} = {{\rm{E}}_{{\varphi _e}}}({{\bf{\hat y}}_d}),
\end{equation}
where \({{\rm{E}}_{{\varphi _e}}}( \cdot )\) denotes a function with learnable parameter \({\varphi _e}\) for embedding the diagnostic decision. This embedded vector refines the visual feature with the channel attention as
\begin{equation}
    {{\bf{f}}_{embed}}(n,m,k) = {{\bf{f}}_v}(n,m,k) \cdot \alpha _d^{embed}(k),
\end{equation}
where \({{\bf{f}}_{embed}}\) denotes a diagnosis embedded feature and \({\alpha _d}^{embed}(k)\) is the k-th element of \({\alpha _d}^{embed}\). From the diagnosis embedded feature, the visual justification \({\alpha _d}^{vis}\) is generated as followings: 
\begin{equation}
    {\alpha _{va}}(n,m) = {{\rm{G}}_{{\varphi _{vis}}}}({{\bf{f}}_{embed}}(n,m,k)),
\end{equation}
\begin{equation}
    {\alpha _d}^{vis}(n,m) = \frac{{\exp ({\alpha _{va}}(n,m))}}{{\sum\limits_n {\sum\limits_m {\exp ({\alpha _{va}}(n,m))} } }},
    \label{eq:softmax}
\end{equation}
where \({\alpha _{va}}(n,m)\) is the 2D map obtained by a function \({{\rm{G}}_{{\varphi _{vis}}}}( \cdot )\) with learnable parameter \({\varphi _{vis}}\). The softmax operation in Eq. (\ref{eq:softmax}) was conducted to represent more focused areas and suppress the activation on the background. For obtaining the textual justification, a text generating feature \({{\bf{f}}_{text}}\) is encoded from the diagnosis embedded feature \({{\bf{f}}_{embed}}\). The text generating feature is used as the input of the textual justification generator which is designed with the Long Short Term Memory(LSTM) networks. The text generating feature \({{\bf{f}}_{text}}\) is obtained by
\begin{equation}
\begin{split}
    &{{\bf{f}}_{embed + vis}}(n,m,k) \\&={{\bf{f}}_{embed}}(n,m,k) \cdot {\alpha _d}^{vis}(n,m) \cdot {\alpha _d}^{embed}(k),
\end{split}
\end{equation}
\begin{equation}
    {{\bf{f}}_{text}} = {{\rm{T}}_{{\varphi _{test}}}}{\rm{(}}{{\bf{f}}_{embed + vis}} + {{\bf{f}}_{embed}}),
\end{equation}
where \({{\bf{f}}_{embed + vis}}\) denotes the refined diagnosis embedded feature by the spatial attention as \({\alpha _d}^{vis}\) and the channel-wise attention as \({\alpha _d}^{embed}\). \({{\rm{T}}_{{\varphi _{test}}}}{\rm{(}} \cdot )\) is a function with learnable parameter \({\varphi _{test}}\) for encoding the text generating feature.\({{\rm{G}}_{{\varphi _{vis}}}}( \cdot )\) and \({{\rm{T}}_{{\varphi _{test}}}}{\rm{(}} \cdot )\) are implemented by multiple convolutional layers. Finally, the textual justification \({\bf{W}} = [{w_1},{w_2}, \cdots ]\) is generated by using the two-hidden-layer-stacked LSTM network \({f^{LSTM}}( \cdot )\) as 
\begin{equation}
    {h_t} = {f^{LSTM}}({{\bf{f}}_{text}},{w_{t - 1}},{h_{t - 1}}),
\end{equation}
\begin{equation}
    {w_t} = {f^{pred}}({h_t}) = {\rm{softmax(}}{W_{pred}}{h_t}{\rm{ + }}{b_{pred}}{\rm{)}},
\end{equation}
where \({w_t}\) denotes a t-th word obtained by converting the t-th hidden state \({h_t}\) using a function \({f^{pred}}( \cdot )\) with learnable parameters \({W_{pred}}\) and \({b_{pred}}\).

\vspace*{\secupmargin}
\subsection{Network training using visual word constraint}
In the training stage, the textual difference loss \({L_D}\) is defined as
\begin{equation}
    \mathcal{L}_D = \sum\limits_{t = 1}^{{l_{text}}} {{\mathop{\rm \text{cross-entropy}}\nolimits}({w_t},w_t^{GT})},
\end{equation}
where textual justification ground truth is \linebreak \({{\bf{W}}^{GT}} = [{w_1}^{GT},{w_2}^{GT}, \cdots ,{w_{{l_{text}}}}^{GT}]\) and \({l_{text}}\) denotes the number of words in ground truth sentence. 
In order to overcome the aforementioned duplication problem in the textual justification generation, we devise a visual word constraint model \({V_{con}}( \cdot )\). The visual word constraint model is designed as a sentence classifier \cite{kim2014convolutional} which predicts the margin and the shape from the given sentences. The margin and the shape are estimated from the given sentences \({\bf{W}}\) as
\begin{equation}
    {{\bf{\hat y}}_{con}} = \{ {{\bf{\hat y}}_{ma}},{{\bf{\hat y}}_{sh}}\}  = {V_{con}}({\bf{W}}),
\end{equation}
where \({{\bf{\hat y}}_{ma}},{{\bf{\hat y}}_{sh}}\) are a predicted margin and a predicted shape, respectively. \({V_{con}}( \cdot )\) denotes a function for predicting the margin and the shape. The visual word constraint model is pre-trained on the training set and utilized to guide the textual justification generator with a visual word constraint loss \({\mathcal{L}_C}\). The visual word constraint loss is defined as
\begin{equation}
\mathcal{L}_{ma} = {\mathop{\rm \text{cross-entropy}}\nolimits}  ({{{\bf{\hat y}}}_{ma}},{\bf{y}}_{ma}^{GT}),
\end{equation}
\begin{equation}
{\mathcal{L}_{sh}} = {\mathop{\rm \text{cross-entropy}}\nolimits}({{{\bf{\hat y}}}_{sh}},{\bf{y}}_{sh}^{GT}),
\end{equation}
\begin{equation}
    {\mathcal{L}_C} = {\mathcal{L}_{ma}} + {\mathcal{L}_{sh}},
\end{equation}
where \({\bf{y}}_{ma}^{GT},{\bf{y}}_{sh}^{GT}\) are ground truth of margin and shape. As a result, overall network is trained by minimizing following loss function:
\begin{equation}
    \mathcal{L} = {\mathcal{L}_D} + \alpha {\mathcal{L}_C},
\end{equation}
where \(\alpha \) is a balancing hyper-parameter. 
By introducing visual word constraint loss, the textual justification could contain more various word. The proposed model could grasp similarity in meaning with word describing same margin or shape even without additional large word set embedding to vector space. 

\vspace*{\secupmargin}
\vspace*{\secupmargin}

\section{Experiments}

\label{sec:experiments}

\vspace*{\secupmargin}



\subsection{Experimental condition}

In the experiments, we used two mammogram datasets. First dataset was the public mammogram dataset, named Digital Database for Screening Mammography (DDSM) dataset \cite{heath2000digital}. The BI-RADS descriptions and the location of masses were annotated by the radiologist \cite{heath2000digital}. The dataset (605 masses) was split into a training set (484 masses) and a test set (121 masses). Second dataset was Full-Field Digital Mammogram (FFDM) dataset from a hospital. A total of 147 masses of 67 patients were collected and a two-fold cross-validation was conducted in this study. The deep network learned from the DDSM dataset was used as the initial network for training of FFDM dataset. The sentence datasets were collected on both the DDSM dataset and the FFDM dataset. Before composing sentences, we investigated words and phrases for describing BI-RADS mass lexicons (margin and shape) in the medical papers \cite{birads, lee2017practical, moon2013quantitative,thomassin2014standardized, selvi_2015, lee2018bi, berment2014masses, surendiran2012mammogram} and its synonyms called visual words. Visual words of each lexicon included 5-12 words or phrases. Three sentences were annotated for each ROI mass image. As shown in Fig. \ref{fig:dataexample}, each sentence was annotated by containing at least one visual word for mass margin and shape respectively. According to \cite{huk2018multimodal}, every sentence included at least 10 words and did not contain BI-RADS mass lexicon as it is. In addition, the sentences contained individual details. 

In order to increase the number of training data, data augmentation was conducted. The two sizes of patches were cropped from the original ROI image at five locations (top left, top right, center, bottom left, bottom right). Each cropped image was also flipped and rotated (0\textdegree, 90\textdegree, 180\textdegree, and 270\textdegree). The size of mini-batch was set to 64 and an Adam optimizer \cite{kingma2014adam} was used with learning rate 0.0005. The balancing parameter was empirically set to 2. 

For the diagnosis network at the front part of the proposed network, we used VGG16 \cite{simonyan15} based binary classifier. The initial weights were pre-trained via ImageNet \cite{deng2009imagenet} and the fine-tuning was conducted. As the visual feature, feature map after conv 5\_3 in the VGG16 network was used in this study. The area under the ROC curve (AUC) was calculated for evaluating the diagnostic performance and the AUC of 0.918 was obtained on the DDSM dataset. During training of the justification generator, the parameters of the diagnosis network were fixed.

\vspace*{\secupmargin}
\vspace*{-1mm}
\subsection{Results}
To validate the effect of our model, we compared the proposed method with the method learned without visual word constraint loss \(\mathcal{L}_C \). Fig. \ref{fig:result1} shows the examples of the visual justification and the textual justification in the proposed method. As shown in the figure, the proposed method could provide the textual justification and the visual justification on the diagnostic decision. The sentences generated by the method learned without visual word constraint loss were also compared. The visual constraint loss \(\mathcal{L}_C\) enabled the textual justification generator to match the margin and shape labels of the generated texture justification and input ROI mass image in the training phase. Therefore, the generated textual justification was more accurate in the proposed method compared to the method without \(\mathcal{L}_C\).

For quantitatively evaluating the quality of the textual justification, we adopted BLEU \cite{papineni2002bleu}, ROUGE-L \cite{lin2004rouge}, and CIDEr \cite{vedantam2015cider} metrics which calculated the similarity between the generated sentence and the reference (ground truth) sentence. Table \ref{table:ddsmmetric} shows the results of the evaluation for the textual justification on the DDSM dataset in terms of BLEU, ROUGE-L, and CIDEr. As shown in the table, with the proposed method learning the model utilizing the visual word constraint loss, the generated textual justifications were closer to reference sentences composed by human. Furthermore, following the evaluations in \cite{Wang_2017_CVPR}, the ratio of the unique sentences and the ratio of the novel sentences were calculated in Table \ref{table:ddsmratio} on the DDSM dataset. The unique sentence was defined as the sentence which was not repeated in all generated sentences and the novel sentence was defined as the sentence which was unseen in the training set. These two metrics were calculated to evaluate the textual justification regarding the duplication problem. If duplication occurred, the textual justification could not accurately narrate the given test image. By calculating the ratio of the novel and unique sentences, it was possible to measure how reliably the textual justification was generated according to the given image. As shown in the table, the number of novel sentences was dramatically improved with the proposed method. The number of the unique sentences in proposed method was also increased compared with the method learned without visual word constraint loss. We conducted same evaluation process on the FFDM dataset. As shown in Table \ref{table:ffdmmetric} and Table \ref{table:ffdmratio}, the proposed method achieved higher score to prove that more accurate and diverse textual justifications were generated.

\vspace*{\secupmargin}
\vspace*{\secupmargin}
\section{Conclusion}
\vspace*{\secupmargin}
\label{sec:conclusion}

\begin{table}[t]
	\caption{Evaluation of textual justification on the DDSM dataset.}
	\label{table:ddsmmetric}
	\vspace*{0.5mm}
	\resizebox{0.999\linewidth}{!}{
		\begin{tabular}{L{3cm} C{3cm} C{3cm}}
			\toprule[1pt]
			& Proposed method & Without \(\mathcal{L}_C\) \\
			\midrule
			BLEU-1  & \textbf{0.3870} & 0.3687    \\
			BLEU-2  & \textbf{0.1968} & 0.1742    \\
			BLEU-3  & \textbf{0.1026} & 0.0887    \\
			BLEU-4  & \textbf{0.0586} & 0.0490    \\
			ROUGE\_L  & \textbf{0.2526} & 0.2439    \\
			CIDEr  & \textbf{0.1514} & 0.1469    \\                                                    
			\bottomrule[1pt]
		\end{tabular}
	}
	
	\vspace*{\tablebetweenmargin}
	\caption{Ratios of the unique sentences and the novel sentences on the DDSM dataset.}
	\label{table:ddsmratio}
	\vspace*{0.5mm}
	\resizebox{0.999\linewidth}{!}{
		\begin{tabular}{L{3cm} C{3cm} C{3cm}}
			\toprule[1pt]
			& Proposed method & Without \(\mathcal{L}_C\) \\
			\midrule
			Ratio of unique sentences   &\textbf{93.39\%}  & 64.46\% \\
			\midrule
			Ratio of novel sentences    &\textbf{43.80\%}  & 4.13\% \\
			\bottomrule[1pt]
		\end{tabular}
	}

	\vspace*{\tablebetweenmargin}
	\caption{Evaluation of textual justification on the FFDM dataset.}
	\label{table:ffdmmetric}
	\vspace*{0.5mm}
	\resizebox{0.999\linewidth}{!}{
		\begin{tabular}{L{3cm} C{3cm} C{3cm}}
			\toprule[1pt]
			& Proposed method & Without \(\mathcal{L}_C\) \\
			\midrule
			BLEU-1  & \textbf{0.4070} & 0.3835    \\
			BLEU-2  & \textbf{0.2296} & 0.2133    \\
			BLEU-3  & \textbf{0.1354} & 0.1187    \\
			BLEU-4  & \textbf{0.0871} & 0.0650    \\
			ROUGE\_L  & \textbf{0.2650} & 0.2596    \\
			CIDEr  & \textbf{0.1366} & 0.1185    \\                                                
			\bottomrule[1pt]
		\end{tabular}
	}

	\vspace*{\tablebetweenmargin}
	\caption{Ratios of the unique sentences and the novel sentences on the FFDM dataset.}
	\label{table:ffdmratio}
	\vspace*{0.5mm}
	\resizebox{0.999\linewidth}{!}{
		\begin{tabular}{L{3cm} C{3cm} C{3cm}}
			\toprule[1pt]
			& Proposed method & Without \(\mathcal{L}_C\) \\
			\midrule
			Ratio of unique sentences   &\textbf{54.42\%}  & 11.56\% \\
			\midrule
			Ratio of novel sentences   &\textbf{65.99\%}  & 8.16\% \\
			\bottomrule[1pt]
		\end{tabular}
	}
\end{table}

In this paper, we proposed the novel deep network to provide multimodal justification for the diagnostic decision. The proposed method could explain the reason of the diagnostic decision with the sentence and indicate the important areas on the image. In the case of textual justification generation for medical purpose, the network tended to generate templated result due to the limited number of medical reports. To overcome this problem, the learning method utilizing visual word constraint loss was devised. By the comparative experiments, the effectiveness of the proposed method was verified. The proposed method generated more diverse and accurate textual justifications. These results imply that the proposed method could explain the diagnostic decision of the deep network more persuasively.

\begin{spacing}{0.85}
	\bibliographystyle{IEEEbib}
	\bibliography{refs}
\end{spacing}

\end{document}